*Original Article*

# Attention based Sequence to Sequence Learning for Machine Translation of Low Resourced Indic Languages – A case of Sanskrit to Hindi

Vishvajit Bakarola[1], Jitendra Nasriwala[2]

[1] *Assistant Professor, Chhotubhai Gopalbhai Patel Institute of Technology, Uka Tarsadia   University, Bardoli, Gujarat, India*
[2] *Associate Professor, Babumadhav Institute of Information Technology, Uka Tarsadia University, Bardoli, Gujarat, India*
[1]vishvajit.bakrola@utu.ac.in

**Abstract -** *Deep Learning techniques are powerful in mimicking humans in a particular set of problems. They have achieved a remarkable performance in complex learning tasks. Deep learning inspired Neural Machine Translation (NMT) is a proficient technique that outperforms traditional machine translation. Performing machine-aided translation on Indic languages has always been a challenging task considering their rich and diverse grammar. The neural machine translation has shown quality results compared to the traditional machine translation approaches. The fully automatic machine translation becomes problematic when it comes to low-resourced languages, especially with Sanskrit. This paper presents attention mechanism based neural machine translation by selectively focusing on a particular part of language sentences during translation. The work shows the construction of Sanskrit to Hindi bilingual parallel corpus with nearly 10K samples and having 178,000 tokens. The neural translation model equipped with an attention mechanism has been trained on Sanskrit to Hindi parallel corpus. The approach has shown the significance of attention mechanisms to overcome long-term dependencies, primarily associated with low resources Indic languages. The paper shows the attention plots on testing data to demonstrate the alignment between source and translated words. For the evaluation of the translated sentences, manual score based human evaluation and automatic evaluation metric based techniques have been adopted. The attention mechanism based neural translation has achieved 88% accuracy in human evaluation and a BLEU score of 0.92 on Sanskrit to Hindi translation.*

**Keywords** — *Attention Mechanism, Low-resourced languages, Neural Machine Translation, Sanskrit, Sequence to Sequence Learning*

## I. INTRODUCTION

Humans have several different ways to communicate with each other. Spoken and written languages are among the most preferred communication ways. To bridge the gap between languages, it is essential to convert a foreign language to a regional language, and the process is known as the translation process. The translation is a complicated and time-consuming process that requires grammatical and domain knowledge of both languages. Typically, machine translation is converting input language (source language) to output language (target language), preserving its semantics. Initially, this process was carried out by a human expert, which is accurate enough for a specific domain at a given time. However, human translation is tedious and time-consuming. With a human translator, reliability is the next crucial issue for different experts concerned with the translation process, and the end translation may vary. The first notable step in computer-aided machine translation was taken in the 1950s. Since then, the efforts have focused on developing a fully automatic machine translation system that accurately mimics human-level fluency [1]. The primary research in machine translation is to melt away the language barrier and open up literature, communication, and language understanding with ease for everyone.

Machine translation has always been a challenging is a fascinating task for the Indic languages. Having the highly diverse grammar and being the morphologically reach languages, machine translation on Indic languages still requires tremendous development efforts. The work focused on developing a fully automatic machine translation system keeping Sanskrit as one of the language pairs. Sanskrit is a language of ancient India and is considered as mother of almost all Indo-European languages. Sanskrit and Hindi both belongs to the Indo-Aryan language family. In the linguistic community, Hindi has been regarded as a descendent of classical Sanskrit through Prakrit [1, 2]. In India, 43.63 percent of the total population are native Hindi speakers. The world report shows that nearly 4.5 percent of the world population are Hindi speakers, which is just 0.5 percent less than native English speakers. Sanskrit is the world's oldest natural language written in most scientific ways. Being an existing human spoken language, Sanskrit is one of the official 22 languages of India according to the eight-schedule of India's constitution. In 2009, Sanskrit was declared the second official language of Uttarakhand and Himachal Pradesh's state in India. Being the primary language of ancient times, all four Vedas and six primary fields of study to learn the Vedas had been written in Sanskrit. The considerable literature available in Sanskrit and its inaccessibility due to lack of understanding is the primary motivation of machine translation work on Sanskrit.

The paper presents the work performing Sanskrit to Hindi machine translation with Neural Machine Translation (NMT) approach. The rest of the article is composed as follows. Section 2 discusses the vastness and complexity of Sanskrit grammar. Section 3 presents several distinctive traditional machine translation approaches and work done on Sanskrit based on those approaches. Section 4 unfolds the NMT along with its significance and major types that deliver a human-like translation. Section 5 details the environment setup, followed by Section 6, showing results and evaluation matrices on machine-translated target language sentences. Finally, Section 7 concludes the work with its future perspectives.

## II. LITERATURE REVIEW

The journey of machine translation has begun in the late 1950s. Rule-based machine translation is the oldest and most foundational approach, further divided into transfer and interlingua-based translation. Over time with the increasing demand and availability of digital text data, it has been observed that the evolution of various state-of-art approaches. Example-based machine translation and statistical machine translation are among those that require corpora and are classified broadly under corpus-based methods [9]. The work on machine translation keeping Sanskrit as one of the language pairs started nearly 30 years back. Desika was the first system developed in the year 1992 [10]. This section presents other works carried out on the Sanskrit language.

### A. Statistical Machine Translation

The statistical machine translation model uses statistical models with parameters derived from the analysis of the bilingual corpora. Statistical machine translation is a corpus-based approach, and they do not know linguistic rules. This system is good at fluency and catching exceptions to the rules [7].cIn 2007, the statistical machine translation approach was used for Google translate, which supported English to Sanskrit translation with other 57 world languages [8].

### B. Rule based Machine Translation

The rule-based model generated the translation of a source language using pre-defined and manually designed grammatical rules. The rules-based models are easy to implement, and they occupy comparatively small memory space. One of the significant advantages of this approach is, it does not require sizeable bi-lingual language corpora. However, the design of grammatical rules is a language-dependent, tedious, and highly time-consuming process.

In 2012, a rule-based approach was carried out on English to Sanskrit translation and applied to 20 random English sentences. The author has reported a BLEU score of 0.4204 [5]. In 2015, work was carried out on English to Sanskrit translation using context-free grammar techniques [6]. In 2017, the interlingual machine translation approach was adopted for Sanskrit to English translation [11]. The work has given significant insights for intermediate language representation and used the Paninian system from Karak analysis.

### C. Other Works on Machine Translation that using Sanskrit

Two works have reported using the neural network approach to achieve translation with the Sanskrit language. In 2019, corpus-based machine translation system with a neural network had been developed for Sanskrit to Hindi translation. The author has reported that their system is better than a rule-based system with a 24 percent higher BLEU score and 39.6 percent less word error rate [12]. Another work carried out in 2019 uses a recurrent neural network for sequence-to-sequence translation [13]. In 2020, the augmented translation technique with Zero-Shot Translation was carried out to translate Sanskrit to Hindi. The author has reported a BLEU score of 13.3, with a higher side stemming from pre-processing [20].

## III. NEURAL MACHINE TRANSLATION

Neural Machine Translation or NMT is the most recent approach to achieve automatic machine translation. NMT uses a neural network to model the conditional probability of the target sequence over the source sequence. NMT has an excellent capability to overcome the traditional machine translation models' shortcomings and provide comparatively efficient human-like fluent translation.

Neural networks learn the source sequence and relate it with an appropriate target sequence mimicking the human way to do this process. Recurrent Neural Network or RNN has been considered for this task, as RNNs models the long-term dependencies between sources and the target languages. Usually, RNN suffers from Exploding gradient – is refer to the problem where network faces increase in weights due to explosion of the long-term components, and Vanishing gradient – is direct to the situation where network weight gets updated with a significantly lower rate of change and the network cannot learn over long term components. And this restricts vanilla RNNs from learning long-term dependencies [14].

Recurrent Neural Network or RNN uses two significant variants – Long Short-Term Memory (LSTM) and Gated Recurrent Unit (GRU) [15], especially to overcome the long-term dependencies learning problem of vanilla RNNs.

### A. Encoder-Decoder Model

The Encoder-Decoder model is an extension of the vanilla RNN model, which make use of two dedicated networks for encoding and decoding the language sequences, as shown in Figure 1. RNN are good at mapping the input and output sequences when their alignment is ahead of time. At training, the input sequence pair to the model is provided, and the model predicts the next word until it meets the sequence end markers [16].

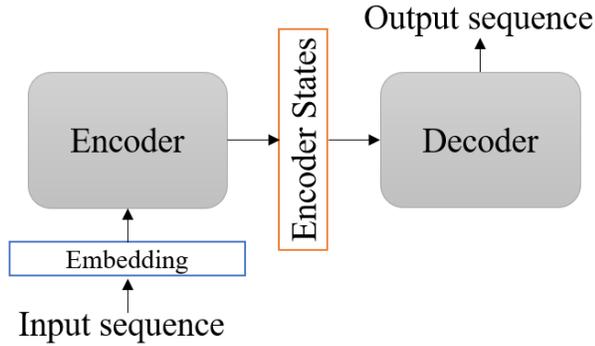

**Fig. 1 Encoder-Decoder Architecture**

The hidden layer represents the actual meaning of the sentence, and it is then fed to the rest of the sequence in the target language. This process gets repeated until the acceptable translation is achieved. Let X be the source language and Y be the target language. The encoder network converts the source sentence $x_1, x_2, x_3, ..., x_n$, into fixed dimension vector space. The decoder network's task is to predict one word at a time with conditional probability as Eq. 1.

$$P(Y|X) = P(Y|X_1, X_2, X_3, ..., X_k) \qquad (1)$$

In Eq. 1, the given sequence $X_1, X_2, X_3, ..., X_k$ does the encoder network encode the fixed dimension vector. Each term used in the distribution will be represented by the softmax layer, i.e., the last layer of the network, which ultimately returns to each class label's probability.

The LSTM learns the conditional probability $P(y_1, ..., y_{T'}|x_i, ..., x_T)$. Here, $x_i, ..., x_T$ is input sequence with its corresponding output sequence $y_1, ..., y_{T'}$, whose length T' may vary from T.

After feeding the input sequence to the LSTM, the hidden state of the LSTM contains the sequence embedding. Finally, this representation is provided to output LSTM having the hidden states $v$. Eq. 2 shows the calculation of the probability for the output sequence.

$$P(y_1, ..., y_{T'}|x_i, ..., x_T) = \prod_{t=1}^{T} P(y_t|v, y_1, ..., y_{t-1}) \qquad (2)$$

## B. Sequence to Sequence Learning with Attention

In sequence-to-sequence learning, the model collectively memorizes the whole vector of the input source sequence, and the decoder uses these encoder states to generate the target sequence. This situation enforces the model to learn small sequences fine, but the model faces trouble learning large sequences, often encountered in language translation problems. One solution to overcome this and continue learning long sentences, even with more than 50 words, focuses on the source sequence's selective part [17]. Fundamentally, to overcome this problem, instead of encoding the whole sequence in a single vector, it is preferable to encode each word of the sequence into a vector [18] and use these vectors in the process of decoding. With this approach, the small sequences have a small length vector, and large sequences have a significant vector since the total number of words in the given sequence is equal to the number of vectors.

It has been observed from the previous encoder-decoder architecture that the encoder results in a given sequence at the end of the entire process. The decoder is forced to find the relevant translation with the use of encoder representation. This ultimately shows that the decoder requires every piece of information to perform the translation. However, this is not the problem with the more minor sequences, but it becomes hard to decode the entire sequence from a single vector as the sequence size increases. The attention mechanism is a way forward. In practice, with natural languages, it is not always suggested to look at the state immediately preceding the present state. Instead of this, some other conditions need to be looked at by the decoder. The foundational idea behind the attention mechanism is that the decoder network's output depends on the weightage combination of all the input sequence states rather than only the immediately previous one [17, 18].

The new architecture focusing on the attention mechanism was proposed in 2015, resolving long-term dependencies with LSTM. The architecture consists of bidirectional RNN as an encoder and decoder that simulates searching through the input sequence during decoding [18]. The goal is to maximize the conditional probability of the target sequence given the source sequence. In the model, each conditional probability will be defined as Eq. 3.

$$P(y_i|y_1, ..., y_{i-1}, X) = g(y_{i-1}, s_i, c_i) \qquad (3)$$

Here, $s_i$ is hidden state of RNN for time i, which is further computed with Eq. 4. The context vector $c_i$ is similar to the vector v presented in Eq. 2.

$$s_i = f(s_{i-1}, y_{i-1}, c_i) \qquad (4)$$

The context vector $c_i$ is depends on sequences of annotations to which the decoder maps the input sequence. The $c_i$ is computed with Eq. 5.

$$c_i = \sum_{j=1}^{T_x} \alpha_{ij} h_j \qquad (5)$$

Where, the $\alpha_{ij}$ is a weight vector and it is computed for each annotation $h_j$ as Eq. 6.

$$\alpha_{ij} = \frac{\exp(e_{ij})}{\sum_{k=1}^{T_x} \exp(e_{ik})} \text{ and } e_{ij} = a(s_{i-1}, h_j) \qquad (6)$$

This alignment model shows how well the inputs around position j and the output at position i get matched. The alignment model is represented as a feedforward neural network. In traditional machine translation systems, this alignment is not explicitly modeled. Figure 2 depicts the functional architecture of the alignment model from [18].

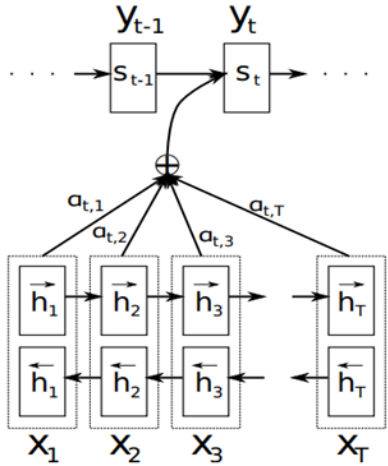

**Fig. 2 The architecture of model trying to generate T-th target word $y_T$ when fed with the input sequence $x_1, \ldots, x_T$ [18]**

## IV. EXPERIMENT SETUP

### A. Dataset

The bilingual corpus of Sanskrit to Hindi language pairs has been developed. The corpus contains 10K Sanskrit sentences parallel translated into Hindi sentences, as shown in Table 1. The Sanskrit sentences are obtained majorly focusing on real-life events from online and offline resources. Help from the linguist community and Sanskrit scholars have been taken to develop and validate human translation.

**TABLE 1. Statistics of Sanskrit-Hindi Bilingual Corpus**

| Language Pair | Samples | Tokens |
|---|---|---|
| Sanskrit | 10650 | 76674 |
| Hindi | 10650 | 101690 |

### B. System Environment Setup

The sequence-to-sequence machine translation model based on Bahdanau's attention [18] has been trained with Sanskrit to Hindi bilingual dataset. The model is designed with 1024 embedding dimensions and Adam as an optimizer [19]. Further, the hyperparameters are tuned with trial-and-error methods. The model is trained with early stopping criteria on Tesla T4 GPUs with 16 GBs of memory.

### C. Data Pre-processing

The present work is using the Sa-Hi language pair from the dataset shown in Table 1. The spell normalization is a significant issue in data pre-processing with the Devanagari script. In Hindi text normalization, words with Persio-Arabic origin are specially taken care of in order to preserve the actual semantics. As the data encoded in Unicode has more than one way of storage, all words have been represented in the same way for normalization. Further, the pre-processing of numbers and the named entity has been carried out to establish uniformity in the corpus.

## V. RESULTS AND EVALUATION

The model was tested for more than a hundred sentences of source Sanskrit language. The evaluation of the target Hindi language was carried out through two different approaches. The first approach works on score based human evaluation. In this approach, four different scores have been proposed as shown in Table 2. The score based human evaluation approach is used for manual verification of model generated target language sentences. Here, human linguist has evaluated target sentences given the source sentences on the scale of 4, Where score 4 represents the completely correct sentence in both syntactic and semantic aspects and the score 1 represents that the sentence is wrong is both syntactic and semantic aspects and delivering no meaning given the source sentence.

In the second approach, an automatic evaluation of target language with BLEU metric [21] has been followed. BLEU score is a widely used metric that is use to calculate the accuracy of model generated sentences in comparison to reference sentences by human linguist in the target language. The BLEU score has been considered in the range of 0 to 1.

**TABLE 2. The Score based Human Evaluation**

| Score | Meaning |
|-------|---------|
| 4 | The translation is completely correct in both syntactic and semantic aspects. |
| 3 | The translation is not entirely correct, but it represents the partial semantic meaning of the source sentence. |
| 2 | The translation is syntactically correct but makes no sense in favor of the source sentence. |
| 1 | The translation is incorrect in both syntactic and semantic manner. |

In testing, the model has obtained an accuracy of 88% with score based human evaluation method and a BLEU score of 0.92. However, coming across a new vocabulary, the model is generating both semantically incomplete sentences. Several sentences from test data are shown in Appendix B. The attention plots have been presented for selected sentences, which are also part of results shown in Table 3 results. It has been observed that the model delivers strong attention between words having a more significant frequency of occurrence with verities of correlation. The attention plots on several results are shown in Appendix A.

The Indic machine translation system has been deployed locally with a user-friendly web interface by integrating the neural machine translation model in the backend, as shown in Fig. 3.

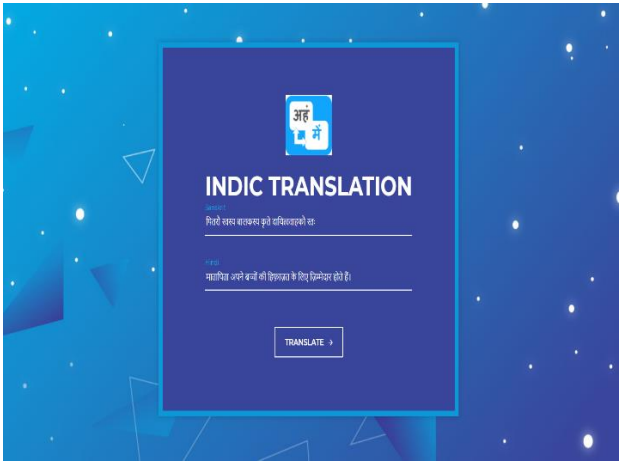

**Fig. 3 Indic Machine Translation System Interface**

## VI. CONCLUSION

The work shows the significance of the attention mechanism to overcome long-term dependencies associated with the vanilla LSTM model during sequence-to-sequence learning. Being the low-resourced language, significantly less amount of digital content is available for Sanskrit.

Considering this as the challenge, The Sanskrit to Hindi bilingual parallel corpus has bene constructed with more than 10K samples and 178,000 tokens. The corpus has been developed in association with the linguist community and used for training the neural machine translation model after required pre-processing and validation.

The LSTM based sequence-to-sequence model has been trained with Bahdanau's attention on the parallel corpus. It has been observed from the experimentation that the model performs well by focusing only on the relevant portion of information in the sentence. After sufficient training with the proper tuning of hyperparameters, the model gives the human evaluation accuracy of 88% and a BLEU score of 0.92 on the unseen Sanskrit sentences. From Table 3, it has been observed that the results are not meeting the appropriate expectations for few sentences as the model is coming across the new vocabulary. The attention plots demonstrate the alignment between the source and target words.

## APPENDIX A

### TABLE 4. Attention Plots of Sample Translations

Source: \<start\> अहं बहु व्यस्तः अस्मि \<end\>
Target: मैं बहुत व्यस्त हूँ \<end\>

Source: \<start\> अहं एकाकिनी अस्मि \<end\>
Target: मैं अकेली हूँ \<end\>

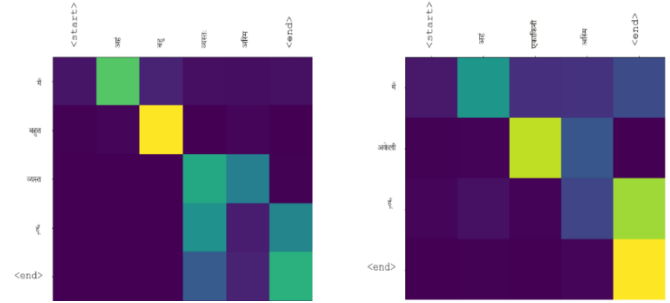

Source: \<start\> अहं तर्तुं शक्नोमि \<end\>
Target: मैं तैर सकता हूँ \<end\>

Source: \<start\> अन्तः आगनुं शक्नोमि? \<end\>
Target: अंदर आ सकता हूँ क्या? \<end\>

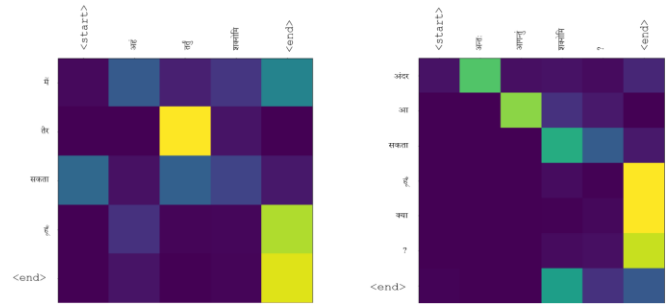

## APPENDIX B

## TABLE 3. Sample Translation through the System

| Source | अन्तः आगन्तुं शक्नोमि ? |
|---|---|
| Reference | अंदर आ सकता हूँ क्या ? |
| Translated | अंदर आ सकता हूँ क्या ? |
| Source | अहं तर्तुं शक्नोमि |
| Reference | मैं तैर सकता हूँ |
| Translated | मैं तैर सकता हूँ |
| Source | अहं एकाकिनी अस्मि |
| Reference | मैं अकेली हूँ |
| Translated | मैं अकेली हूँ |
| Source | पितरी स्वस्य बालकस्य कृते दायित्ववाहकी स्तः |
| Reference | मातापिता अपने बच्चों की हिफाज़त के लिए ज़िम्मेदार होते हैं |
| Translated | मातापिता अपने बच्चों की हिफाज़त के लिए ज़िम्मेदार होते हैं |
| Source | जापानदेशः विश्वस्य देशेषु एकः अर्थतंत्रः देशः अस्ति |
| Reference | जापान दुनिया के सबसे ताकतशाली अर्थतंत्रों में से एक है |
| Translated | जापान दुनिया के सबसे ताकतशाली अर्थतंत्रों में से एक है |
| Source | प्रवेशात् पूर्वं पादका त्याज्या |
| Reference | अपने हाथ में मरना उसके बाहर जाने की कोशिश करो |
| Translated | अपने हाथ में मरना उसके बाहर जाने की कोशिश करो |
| Source | अहं बहु व्यस्तः अस्मि |
| Reference | मैं बहुत व्यस्त हूँ |
| Translated | मैं बहुत व्यस्त हूँ |

## ACKNOWLEDGEMENT

We would express our gratitude to the Indic linguist community. Their work has helped us to retrieve insights into both Sanskrit and Hindi grammar. We would like to acknowledge Shri Udit Sharma and Shri Harshad Joshi, who help us construct and validate our parallel corpus. We are grateful to everyone who has directly or indirectly proven helpful in our work. We are also thankful to other researchers whose work helps us derive some conclusions and identify the problems.

## REFERENCES

[1] D. Jitendra Nasriwala and V. Bakarola, Computational Representation of Paninian Rules of Sanskrit Grammar for Dictionary-Independent Machine Translation, vol. 1046, no. July. Springer Singapore, (2019).

[2] A. C. Woolner, "Introduction to Prakrit.pdf." University of the Panjab, Lahore, (1917).

[3] P. Kiparsky, "On the Architecture of Panini's Grammar," in Sanskrit Computational Linguistics: First and Second International Symposia Rocquencourt, France, October 29-31, 2007 Providence, RI, USA, May 15-17, 2008 Revised Selected and Invited Papers, Berlin, Heidelberg: Springer-Verlag, (2009) 33–94.

[4] B. Panchal, V. Bakrola, and D. Dabhi, "An Efficient Approach of Knowledge Representation Using Paninian Rules of Sanskrit Grammar BT - Recent Findings in Intelligent Computing Techniques," (2018) 199–206.

[5] V. Mishra and R. B. Mishra, "English to Sanskrit Machine Translation System: A Rule-Based Approach," Int. J. Adv. Intell. Paradigm., vol. 4, no. 2 (2012) doi: 10.1504/IJAIP.2012.048144, 168–184.

[6] P. Bahadur, A. Jain, and D. S. Chauhan, "Architecture of English to Sanskrit machine translation," IntelliSys 2015 - Proc. 2015 SAI Intell. Syst. Conf. (2015) doi: 10.1109/IntelliSys.2015.7361204, 616-624.

[7] P. Koehn, Statistical Machine Translation. Cambridge University Press, (2010).

[8] P. D. Mane and A. Hirve, "Study of Various Approaches in Machine Translation for Sanskrit Language," vol. 2, (2013), 383–387.

[9] U. S. T. Tanveer Siddiqui, Natural Language Processing and Information Retrieval. Oxford University Press, (2015).

[10] P. R. V. Veda, "Computer Processing of Sanskrit," C-DAC, Pune, (1992).

[11] H. S. Sreedeepa and S. M. Idicula, "Interlingua based Sanskrit-English machine translation," Proc. IEEE Int. Conf. Circuit, Power Comput. Technol. ICCPCT, (2017) doi: 10.1109/ICCPCT.2017.8074251.

[12] M. Singh, R. Kumar, and I. Chana, "Corpus based Machine Translation System with Deep Neural Corpus based Machine Translation System with Deep Neural Network for Sanskrit to Hindi Translation Network for Sanskrit to Hindi Translation," Procedia Comput. Sci., vol. 167, (2020), doi: 10.1016/j.procs.2020.03.306, 2534-2544.

[13] N. Koul and S. S. Manvi, "A proposed model for neural machine translation of Sanskrit into English," Int. J. Inf. Technol., (2019), doi: 10.1007/s41870-019-00340-8.

[14] A. Shewalkar, D. Nyavanandi, and S. A. Ludwig, "Performance Evaluation of Deep Neural Networks Applied to Speech Recognition: RNN, LSTM and GRU," J. Artif. Intell. Soft Comput. Res., vol. 9, no. 4, doi: https://doi.org/10.2478/jaiscr-2019-0006, 235-245.

[15] Y. Hu, A. Huber, and S.-C. Liu, "Overcoming the vanishing gradient problem in plain recurrent networks." (2018) [Online]. Available: https://openreview.net/forum?id=Hyp3i2xRb.

[16] I. Sutskever, O. Vinyals, and Q. V. Le, "Sequence to sequence learning with neural networks," Adv. Neural Inf. Process. Syst., vol. 4, no. January, (2014), 3104–3112.

[17] M. T. Luong, H. Pham, and C. D. Manning, "Effective approaches to attention-based neural machine translation," Conf. Proc. - EMNLP 2015 Conf. Empir. Methods Nat. Lang. Process., (2015), doi: 10.18653/v1/d15-1166, 1412-1421

[18] D. Bahdanau, K. H. Cho, and Y. Bengio, "Neural machine translation by jointly learning to align and translate," 3rd Int. Conf. Learn. Represent. ICLR 2015 - Conf. Track Proc., (2015), 1-15.

[19] D. P. Kingma and J. Ba, "Adam: A Method for Stochastic Optimization." (2017).

[20] Rashi Kumar, Piyush Jha and Vineet Sahula, An Augmented Translation Technique for Low Resource Language Pair: Sanskrit to Hindi Translation. In Proceedings of 2019 2nd International Conference on Algorithms, Computing and Artificial Intelligence (ACAI'19). Sanya, China, (2019), https://doi.org/10.1145/3377713.3377774.

[21] Kishore Papineni, Salim Roukos, Todd Ward and Wei-Jing Zhu, BLEU – A Method for Automatic Evaluation of Machine Translation, Proceedings of the 40th Annual Meeting of the Association for Computational Linguistics (ACL), Philadelphia, July (2002), pp. 311-318.